\documentclass[10pt,twocolumn,letterpaper]{article}

\usepackage[pagenumbers]{cvpr} % To force page numbers, e.g. for an arXiv version

\definecolor{cvprblue}{rgb}{0.21,0.49,0.74}
\usepackage[pagebackref,breaklinks,colorlinks,allcolors=cvprblue]{hyperref}

\newcommand{\methodname}{Art3D\xspace}
\newcommand{\datasetname}{\textit{Flat-2D}\xspace}
\def\paperID{39} % *** Enter the Paper ID here
\def\confName{CVPR}
\def\confYear{2025}

\title{\methodname: Training-Free 3D Generation from Flat-Colored Illustration}

\author{
Xiaoyan Cong\footnotemark[1] \quad
Jiayi Shen\footnotemark[1] \quad
Zekun Li \quad
Rao Fu \quad
Tao Lu \quad
Srinath Sridhar \\
Brown University \
}

\begin{document}
\newcommand{\teaserCaption}{
}
\twocolumn[{
    \renewcommand\twocolumn[1][]{#1}
    \maketitle
    % \vspace{-0.9 cm}
    \centering
    \includegraphics[width=0.95\linewidth]{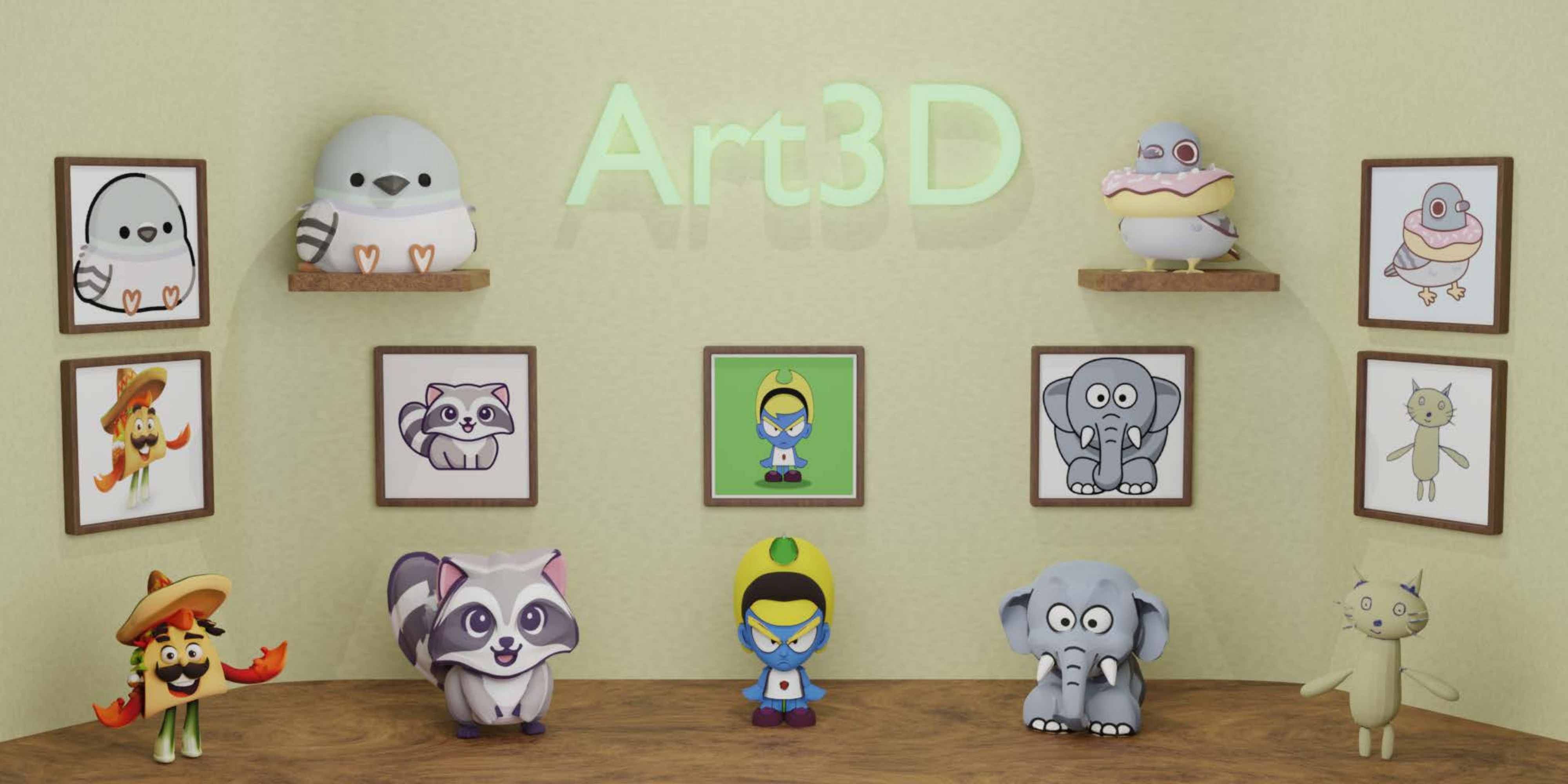}
    % \vspace{-0.7cm}
    \captionsetup{type=figure}
    \captionof{figure}{\teaserCaption Our \methodname creates high-quality 3D assets from a single flat-colored illustration and can be adapted to various drawing styles.}
    \label{fig:teaser}
    \vspace{4mm}
}]

\maketitle
\footnotetext[1]{Equal contribution.}
\begin{abstract}
Large-scale pre-trained image-to-3D generative models have exhibited remarkable capabilities in diverse shape generations.
However, most of them struggle to synthesize plausible 3D assets when the reference image is \textbf{flat-colored} like hand drawings due to the lack of 3D illusion, which are often the most user-friendly input modalities in art content creation.
To this end, we propose \methodname, a training-free method that can lift flat-colored 2D designs into 3D.
By leveraging structural and semantic features with pretrained 2D image generation models and a VLM-based realism evaluation, \methodname successfully enhances the three-dimensional illusion in reference images, thus simplifying the process of generating 3D from 2D, and proves adaptable to a wide range of painting styles.
% By enhancing the three-dimensionality illusion of reference images based on the structure feature through pretrained 2D image generation models, \methodname is generalized to versatile painting styles.
To benchmark the generalization performance of existing image-to-3D models on flat-colored images without 3D feeling, we collect a new dataset, \datasetname, with over 100 samples.
Experimental results demonstrate the performance and robustness of \methodname, exhibiting superior generalizable capacity and promising practical applicability. 
Our source code and dataset will be publicly available on our project page: \href{https://joy-jy11.github.io/}{https://joy-jy11.github.io/}.
\end{abstract}    

\begin{figure}
    \centering
    \includegraphics[width=\linewidth]{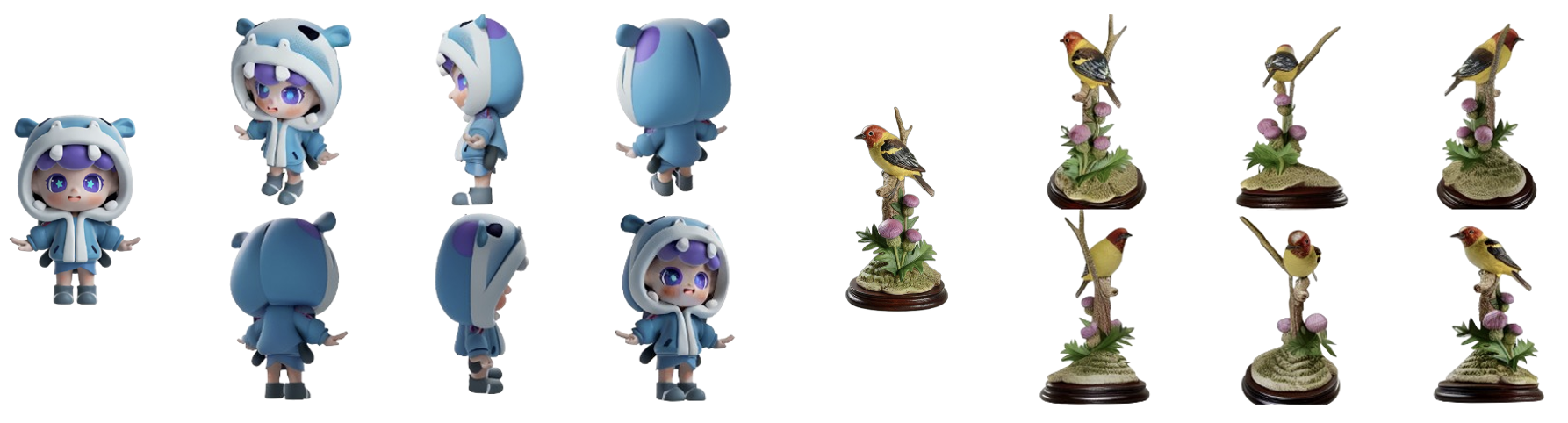}
    \includegraphics[width=\linewidth]{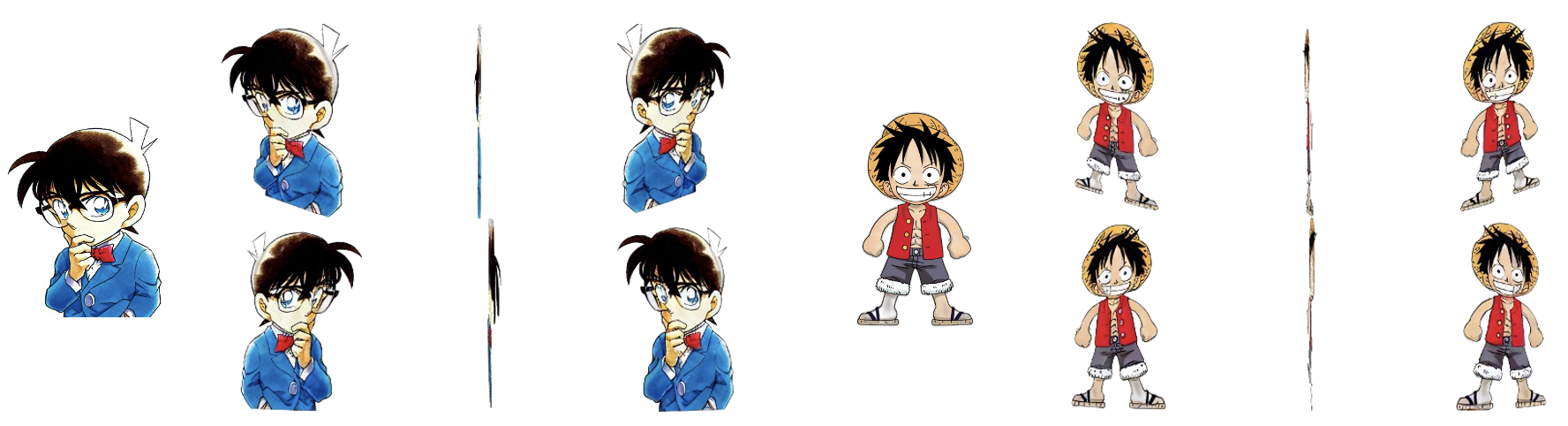}
    \vspace{0.1in}
    \begin{minipage}[t]{0.12\linewidth}
        \centering
        Input
    \end{minipage}%
    \hfill
    \begin{minipage}[t]{0.3\linewidth}
        \centering
        Result
    \end{minipage}%
    \hfill
    \begin{minipage}[t]{0.13\linewidth}
        \centering
        Input
    \end{minipage}%
    \hfill
    \begin{minipage}[t]{0.3\linewidth}
        \centering
        Result
    \end{minipage}
    % \vspace{-0.4cm}
    \caption{We compare the generation results from InstantMesh~\cite{xu2024instantmeshefficient3dmesh} for two kinds of image inputs. 
    The top row presents results obtained from images with a three-dimensional appearance, while the bottom row shows results using flat-colored images as input.
    We observe that while existing methods perform well on data within their training distribution, they tend to degenerate and produce abnormally thin geometric structures when applied to flat-colored images.}

    % \vspace{-0.2in}

    \label{fig:flat_input}
\end{figure}

\section{Introduction}
Generating high-quality 3D assets from single images has widespread applications in virtual reality, video games, filmmaking, \etc.
With the rapid development and application of foundation models~\cite{dhariwal2021diffusionmodelsbeatgans, ho2020denoisingdiffusionprobabilisticmodels, song2020score, Cong_2024_CVPR, zheng2024oscillationinversionunderstandstructure}, recent image-to-3D generative models~\cite{xiang2024structured, nichol2022point,xu2024instantmeshefficient3dmesh,lan2024ln3diff,chen20243dtopiaxlscalinghighquality3d,tang2024lgm} has demonstrated impressive performance on various kinds of image inputs.
However, a common limitation of existing methods shown in \cref{fig:flat_input} is their inability to effectively handle \textbf{flat-colored} images that inherently lack a sense of three-dimensionality, such as hand drawings, cel painting, icons, cartoon-style designs, \etc.
Yet, these types of images are among the most accessible and intuitive tools for users in visual art creation.
This limitation significantly reduces the practical applicability of such methods in real-world creative workflows.

To solve this problem, we address this limitation by introducing \methodname, a training-free 3D generation solution for flat-colored images, enhancing the three-dimensional illusion of the flat inputs to align them with the distribution of training data used in pre-trained image-to-3D models. 
Given a flat input image, \methodname first leverages the strong priors of a pre-trained flow-based ControlNet~\cite{flux2024, zhang2023addingconditionalcontroltexttoimage} to synthesize multiple reference candidates that are implicitly imbued with a 3D appearance, based on their structure features such as depth map and canny edge and semantic information extracted by Visual Language Models (VLM)~\cite{gpt4v, li2023blip}.
Then, we utilize VLM to select the geometry proxy image that is the most realistic and shows the most 3D feeling among candidates for shape generation based on visual question answering. 
With the implicit 3D information in the proxy image, a pre-trained shape generator~\cite{xiang2024structured} can synthesize a reasonable and complete shape consistent with the enhanced flat-colored input.
Finally, we use Hunyuan2.0~\cite{zhao2025hunyuan3d20scalingdiffusion} to bake the texture into the generated mesh based on the original flat-color image.

To validate the generalization of \methodname, we collect a dataset called \textbf{\datasetname}, containing over 100 flat-colored images.
Experimental results demonstrate that, given flat images, \methodname can generate complete and realistic textured meshes, providing a robust and general solution for failure cases where all existing image-to-3D models output abnormal thin structures.
We envision \methodname as a powerful tool for pushing the boundaries of image-to-3D foundation models with broad applicability in filmmaking, art creation, video games, and VR/AR.
\section{Related Work}
% \subsection{Diffusion Models}
% The advent of generative modeling~\cite{bond2021deep, dhariwal2021diffusionmodelsbeatgans, esser2021taming, karras2019style, ramesh2021zero, vahdat2020nvae, zhang2022styleswin} has witnessed rapid progress in past years. 
% Diffusion models~\cite{dhariwal2021diffusionmodelsbeatgans, ho2020denoisingdiffusionprobabilisticmodels, song2020score, yang2023diffusion} have recently shown superior generative ability generalization ability. 
% Diffusion models are a class of generative models that are trained to reverse a Markovian forward process. 
% Given a sample $z_0$ sampled from the data distribution $p(z)$, the forward process generates a series of noised variable $\{ z_t | t \in (0, T)\}$, where $z_t = \alpha_t z_0 + \sigma_t \epsilon$ and $\epsilon$ is a random noise sampled from $\mathcal{N}(\mathbf{0}, \mathbf{1})$. 
% $\{ \alpha_t, \epsilon_t\}$ defines a specific sampling strategy. 
% The noised variable $z_T$ is getting closer to a purely random noise with $T$ growing larger. 
% The diffusion model is trained to recover $z_{t-1}$ from $z_t$ by predicting the added noise during forward process. 
% Diffusion models have achieved remarkable success in 2D generation tasks like text-to-image generation~\cite{gu2022vector, nichol2021glide, ramesh2022hierarchical, rombach2022highresolutionimagesynthesislatent, saharia2022photorealistic}, which serves as a foundation model and enables various appealing applications~\cite{ruiz2023dreambooth, hertz2022prompt, wang2022pretraining, cong2024automatic, zheng2024oscillation}.

\paragraph{Image-to-3D}
Many pioneering works have explored image-conditioned 3D generative modeling on various representations such as point clouds~\cite{melas2023pc2, nichol2022point, tyszkiewicz2023gecco, wu2023sketch, zhou20213d, cong20244drecons4dneuralimplicit}, meshes~\cite{alliegro2023polydiff, liu2023meshdiffusion}, signed distance function fields~\cite{cheng2023sdfusion, chou2023diffusion, shim2023diffusion, zheng2023locally}, and neural fields~\cite{jun2023shapegeneratingconditional3d, gupta20233dgen, muller2023diffrf, wang2023rodin, zhang20233dshape2vecset}. 
% Despite the promising achievement these methods have made, they are hard to have the same generalization capacities as 2D foundation models~\cite{rombach2022highresolutionimagesynthesislatent, yang2024depth, zhang2023addingconditionalcontroltexttoimage} due to the limited scale of high-quality 3D training data.
DreamFusion~\cite{poole2022dreamfusion} generates assets by distilling 2D diffusion priors into 3D representations via a per-scene optimization way. 
% This score distillation sampling (SDS) strategy exhibits superior generalization capacities on the text-to-3D generation task. 
The score distillation sampling based methods~\cite{poole2022dreamfusion, chen2023fantasia3d, Lin_2023_CVPR, Wang_2023_CVPR, wang2023prolificdreamerhighfidelitydiversetextto3d} suffer from time-consuming optimization and the multi-face issue, known as "Janus" problem. 
Instantmesh~\cite{xu2024instantmeshefficient3dmesh} divides generation process into two steps through first synthesizing sparse novel views~\cite{long2023wonder3d, tang2024lgm, liu2023one2345singleimage3d, qian2023magic123imagehighquality3d} and then reconstruction.
TRELLIS~\cite{xiang2024structured} generates 3D objects in an end-to-end way through denoising in a sparse 3D latent space. 
Despite the amazing performance existing image-to-3D methods have presented, all of them fail for flat-colored images shown in Figure~\ref{fig:flat_input}, outputting abnormal thin structures.
\section{Methodology}

\begin{figure*}
    \centering
    \includegraphics[width=0.95\textwidth]{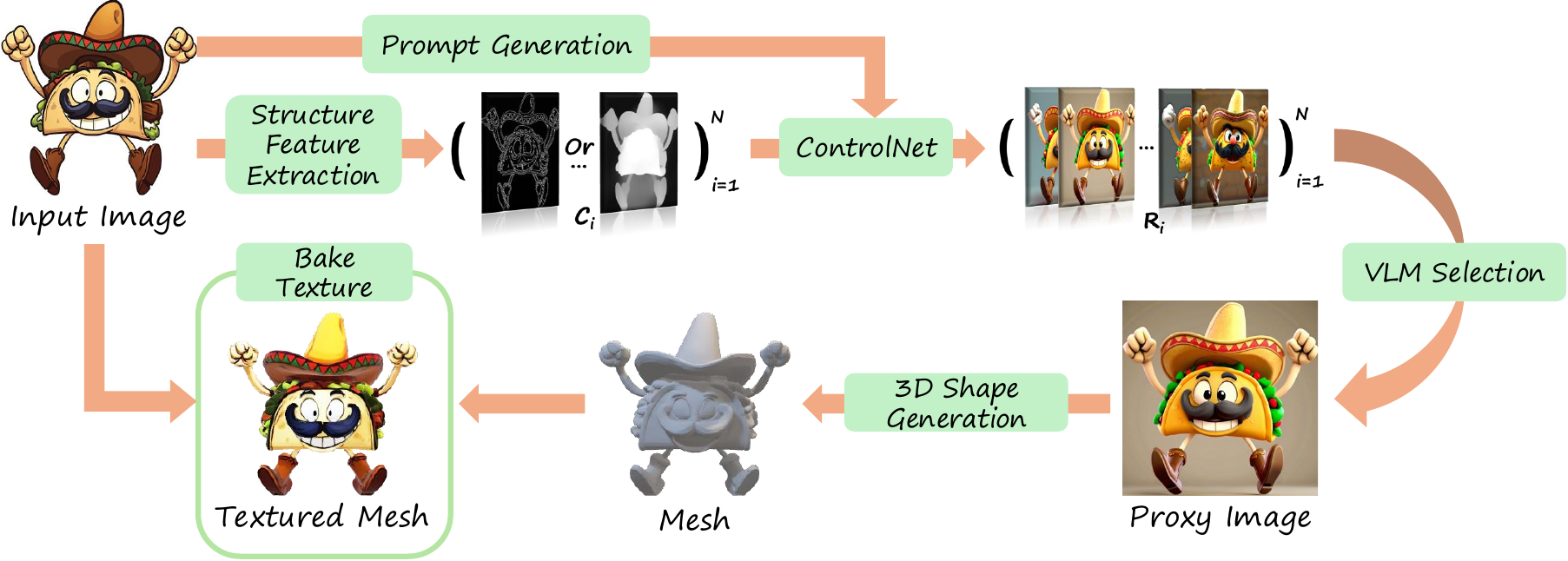}
    % \vspace{-0.1in}
    \caption{\textbf{Pipeline.} \methodname adds 3D illusion to the flat-colored image through ControlNet~\cite{zhang2023addingconditionalcontroltexttoimage} based on the structure features, \eg, canny edge or depth map.
    The mesh is then synthesized based on the proxy image and textured by baking information from the input image.}
    % \vspace{-0.2in}
    \label{fig:pipeline}
\end{figure*}

% \paragraph{Image-to-3D Models.} Feed-forward 3D generative models~\cite{szymanowicz2024splatterimageultrafastsingleview, tang2024dreamgaussiangenerativegaussiansplatting, tang2024lgmlargemultiviewgaussian, zhang2024gaussiancubestructuredexplicitradiance, xu2024instantmeshefficient3dmesh, xiang2024structured} have achieved significant progress in terms of quality, speed- and memory- efficiency. 
% This line of works usually comprise a multi-view diffusion model and a sparse-view large reconstruction model. Given an input image $\mathbf{I}$, pre-trained multi-view diffusion model first synthesizes 3D consistent multi-view images $\{ I_i | i=1,2,\dots\}$, which are fed into pre-trained sparse-view large reconstruction model to generate a high-quality 3D mesh $\mathbb{M}$.

% \subsection{Our Approach}
As shown in Fig.~\ref{fig:flat_input}, existing image-to-3D models fail to generate reasonable geometry from flat-colored inputs, producing an abnormally thin structure.
We attribute this failure mode to the misalignment of data distribution between flat-colored inputs and those used for training image-to-3D models, which contain sufficient 3D cues like shadows, certain viewpoints, and reasonable depth structures, especially those rendered from large-scale realistic 3D object datasets such as Objaverse~\cite{deitke2022objaverseuniverseannotated3d}, OmniObject3D~\cite{wu2023omniobject3d}, and CO3D~\cite{reizenstein2021common}. 
To address this limitation, we propose a plug-and-play augmentation module $\mathcal{O}$ to enhance the three-dimensional illusion of the input flat-colored image $\mathbf{I}$ based on structure features and semantic information through pretrained 2D image diffusion models. 
Leveraging the enhanced proxy image $\hat{\mathbf{I}} = \mathcal{O}(\mathbf{I})$, \methodname can efficiently generate 3D shape by Trellis~\cite{xiang2024structured} and texture the mesh by Hunyuan2.0~\cite{zhao2025hunyuan3d20scalingdiffusion}.

As shown in Fig.~\ref{fig:pipeline}, we first utilize ControlNet~\cite{zhang2023addingconditionalcontroltexttoimage} to control pre-trained flow-based image generation models~\cite{flux2024} to synthesize $N$ reference candidates $\{ \mathbf{R}_i\}_{i=1}^{N}$ given different structure conditions $\{ \mathbf{C}_i\}_{i=1}^{N}$, with the help of captions $\mathbf{T}$ generated by VLM such as GPT-4V(ision)~\cite{gpt4v} and BLIP-2~\cite{li2023blip}: 
\begin{equation}
    \mathbf{R}_i = \mathcal{G}(\mathbf{I}, \mathbf{T}, \mathbf{C}_i).
\end{equation}
Then we select the most satisfactory reference candidate as our proxy image $\hat{\mathbf{I}}$ by leveraging VLM to perform visual question answering. 
Specifically, we provide GPT-4V(ision)~\cite{gpt4v} with all the reference candidates $\{ \mathbf{R}_i\}_{i=1}^{N}$ and ask 'Which image do you think is the most realistic and shows the most 3D feeling?'.
\begin{equation}
    \hat{\mathbf{I}} = \text{VQA}(\{ \mathbf{R}_i\}_{i=1}^{N}).
\end{equation}
Numerous proxy conditions are available, including canny edge~\cite{canny1986computational}, HED boundary~\cite{xie2015holistically}, depth map, semantic segmentation map, \etc.
The overall objective, irrespective of the chosen proxy conditions, is to equip the flat images with a feeling of volume and form.
Empirically, we choose the proxy condition $\mathbf{C}_i$ as the depth map and Canny edge for their ample geometric priors and superior and robust performance.
In our experiments, We generate $N_1$ reference candidates conditioned on the depth map of $\mathbf{I}$ estimated by Midas~\cite{ranftl2020robustmonoculardepthestimation} and $N_2$ reference candidates conditioned on the canny edge $\mathbf{I}$.
Thanks to the strong priors in the pre-trained image generator, our proxy image $\hat{\mathbf{I}}$ contains enough 3D feeling while preserving the identity of the original inputs. 
\captionsetup[subfigure]{labelformat=empty}
\begin{figure*}[h!]
    \centering
    
    \begin{subfigure}[b]{0.122\textwidth}
        \centering
        \includegraphics[width=\textwidth]{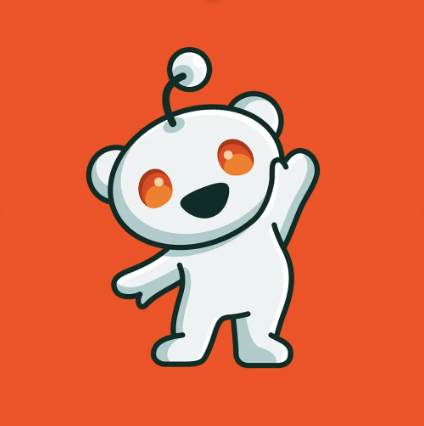}\\
        \includegraphics[width=\textwidth]{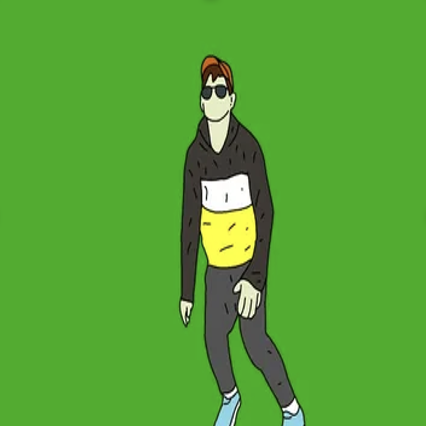}\\
        \includegraphics[width=\textwidth]{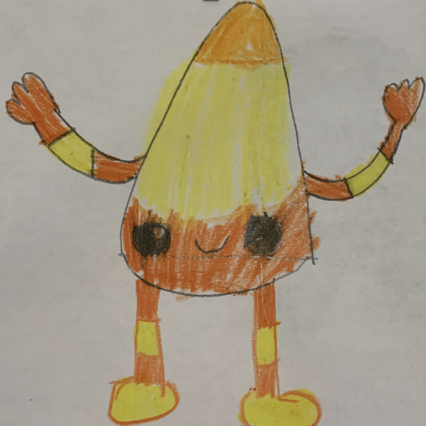}
        \includegraphics[width=\textwidth]{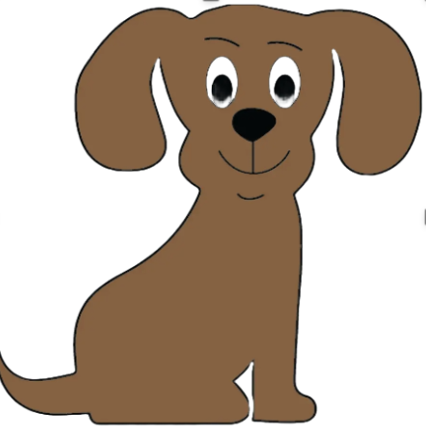}
        \includegraphics[width=\textwidth]{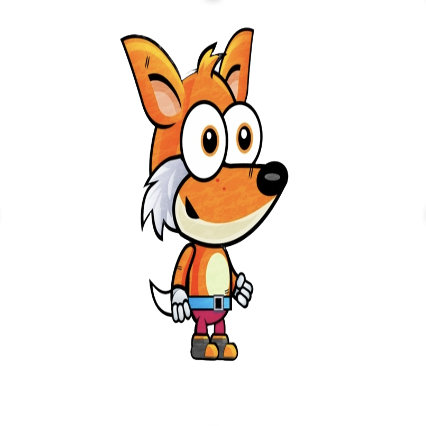}
        \includegraphics[width=\textwidth]{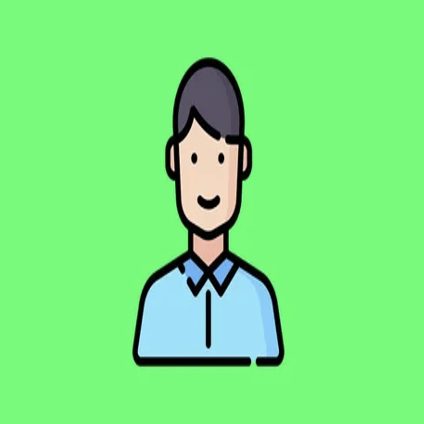}
        \includegraphics[width=\textwidth]{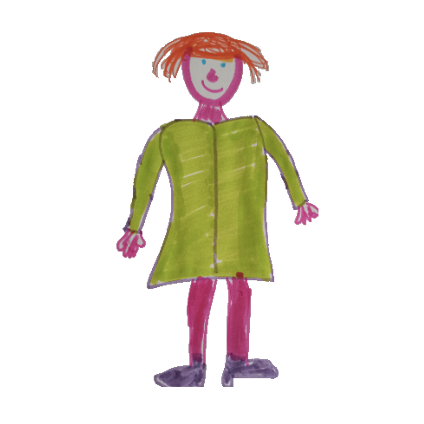}
        \includegraphics[width=\textwidth]{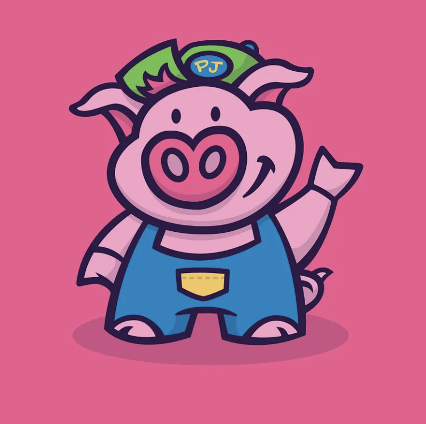}
        \includegraphics[width=\textwidth]{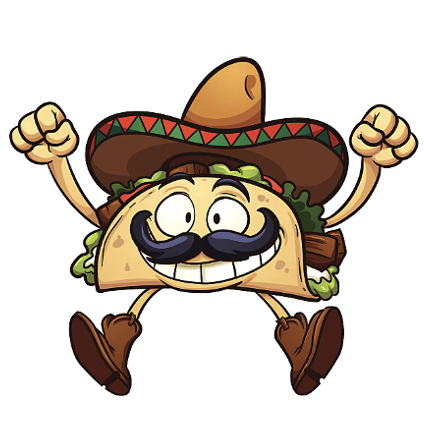}
        \caption{Input Flat Image}
    \end{subfigure}\hspace{-0.1em}
    \begin{subfigure}[b]{0.122\textwidth}
        \centering
        \includegraphics[width=\textwidth]{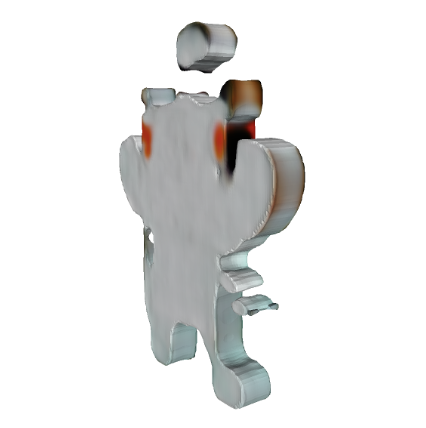}\\
        \includegraphics[width=\textwidth]{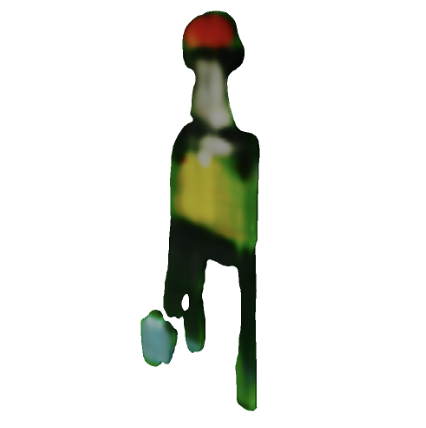}\\
        \includegraphics[width=\textwidth]{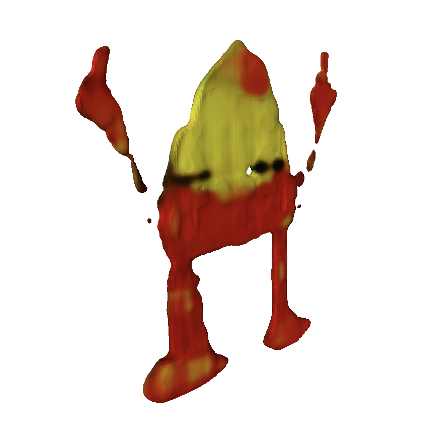}
        \includegraphics[width=\textwidth]{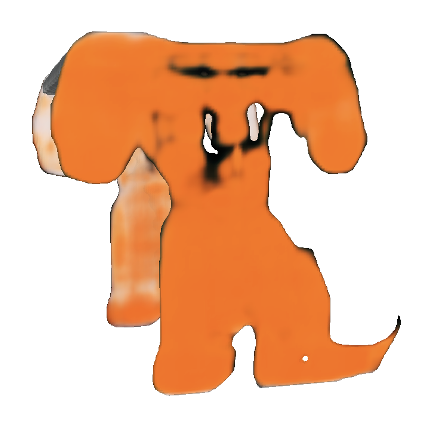}
        \includegraphics[width=\textwidth]{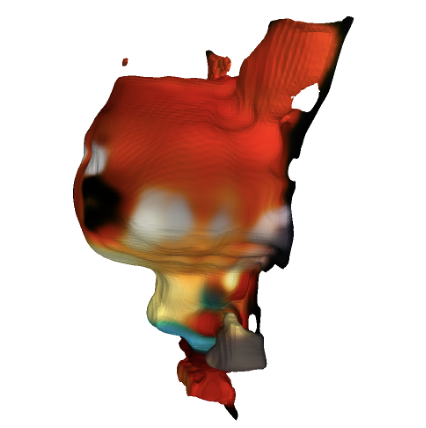}
        \includegraphics[width=\textwidth]{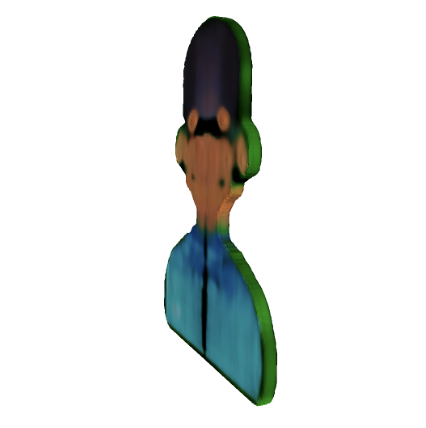}
        \includegraphics[width=\textwidth]{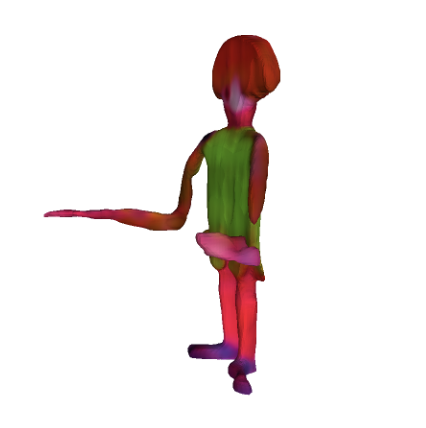}
        \includegraphics[width=\textwidth]{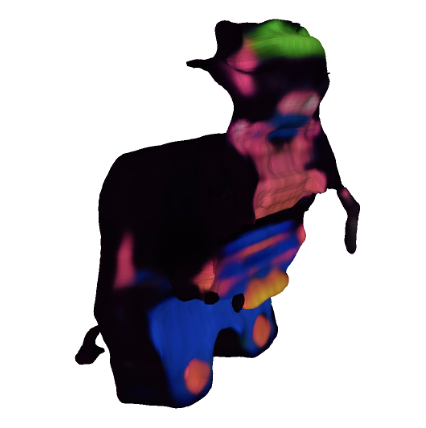}
        \includegraphics[width=\textwidth]{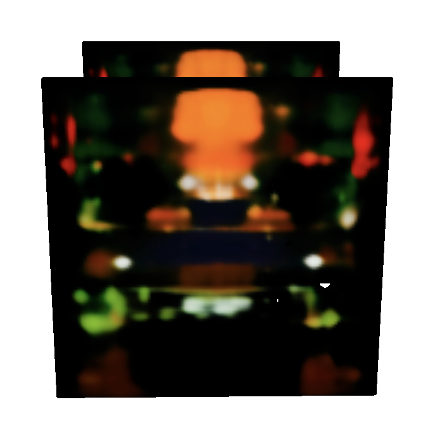}
        \caption{Shap-E~\cite{jun2023shapegeneratingconditional3d}}
    \end{subfigure}\hspace{-0.1em}
    \begin{subfigure}[b]{0.122\textwidth}
        \centering
        \includegraphics[width=\textwidth]{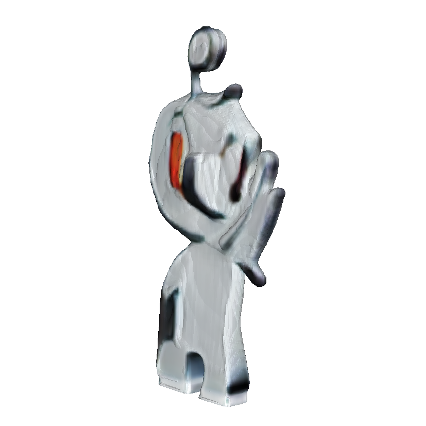}\\
        \includegraphics[width=\textwidth]{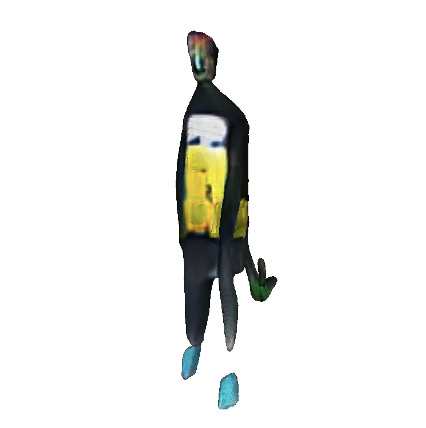}\\
        \includegraphics[width=\textwidth]{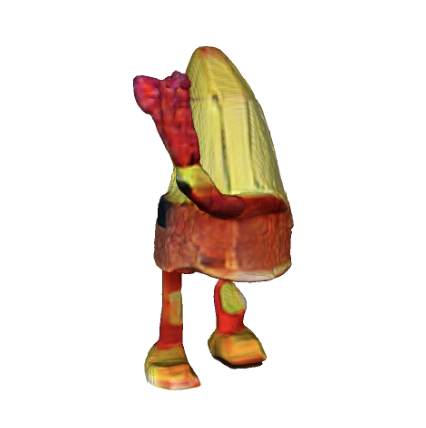}
        \includegraphics[width=\textwidth]{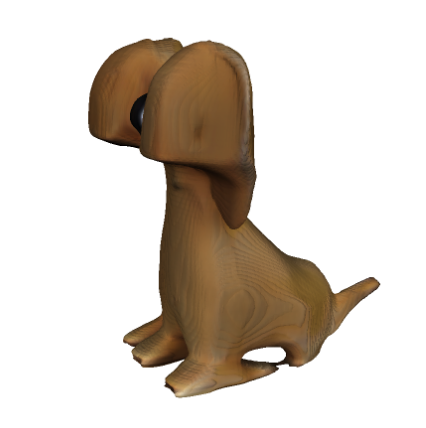}
        \includegraphics[width=\textwidth]{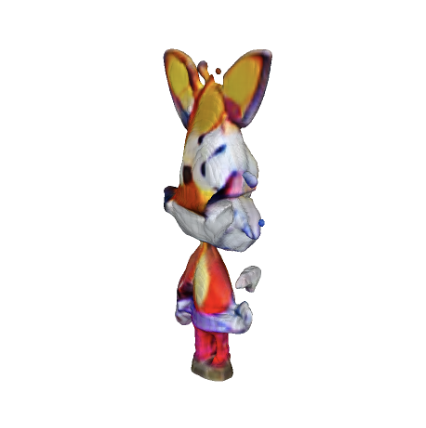}
        \includegraphics[width=\textwidth]{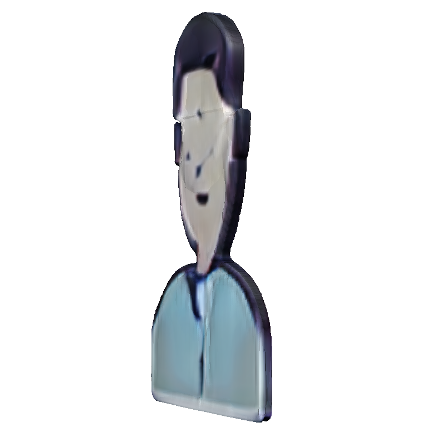}
        \includegraphics[width=\textwidth]{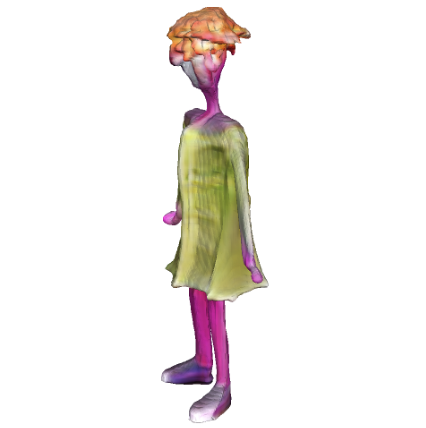}
        \includegraphics[width=\textwidth]{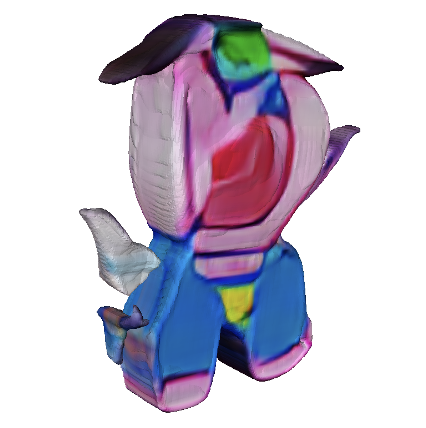}
        \includegraphics[width=\textwidth]{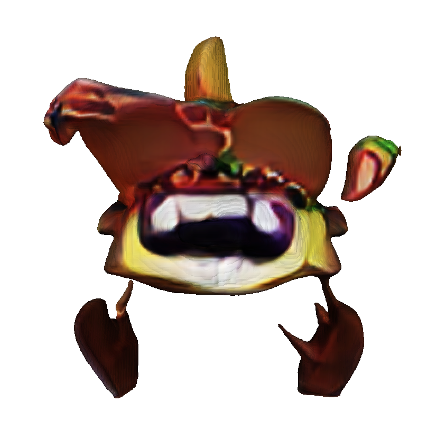}
        \caption{LN3Diff~\cite{lan2024ln3diff}}
    \end{subfigure}\hspace{-0.1em}
    \begin{subfigure}[b]{0.122\textwidth}
        \centering
        \includegraphics[width=\textwidth]{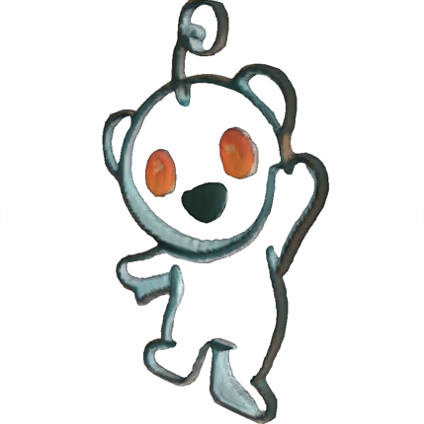}\\
        \includegraphics[width=\textwidth]{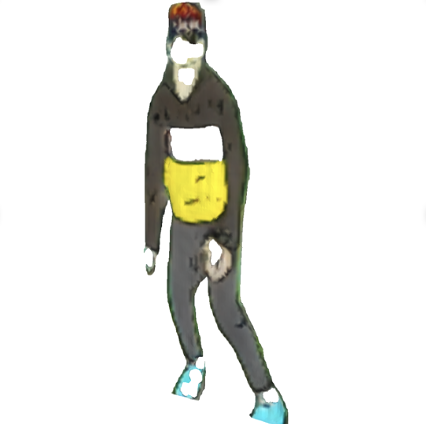}\\
        \includegraphics[width=\textwidth]{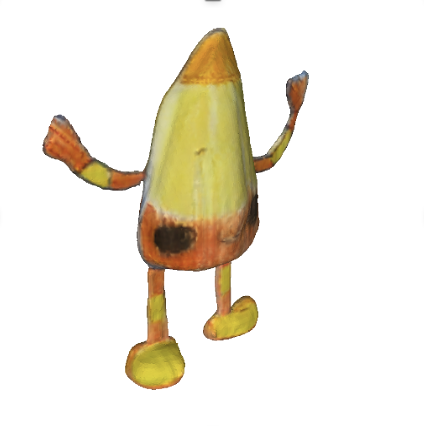}
        \includegraphics[width=\textwidth]{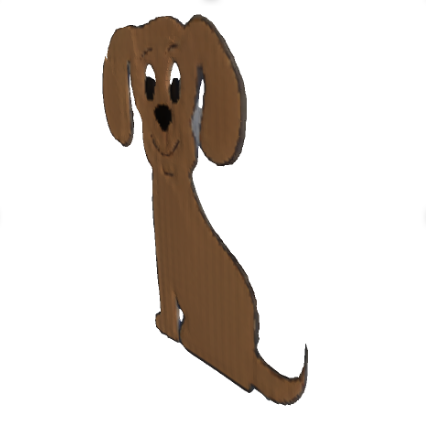}
        \includegraphics[width=\textwidth]{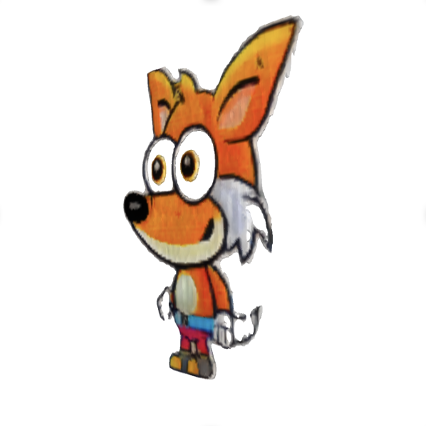}
        \includegraphics[width=\textwidth]{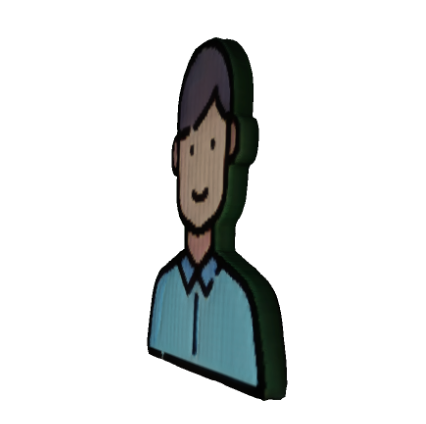}
        \includegraphics[width=\textwidth]{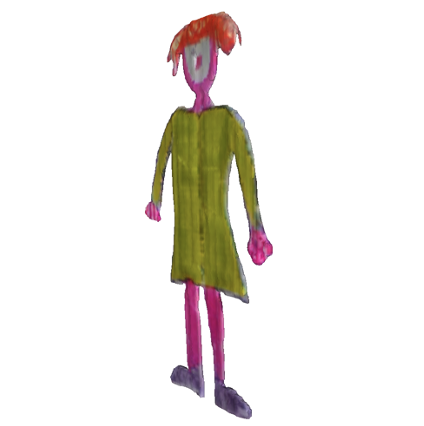}
        \includegraphics[width=\textwidth]{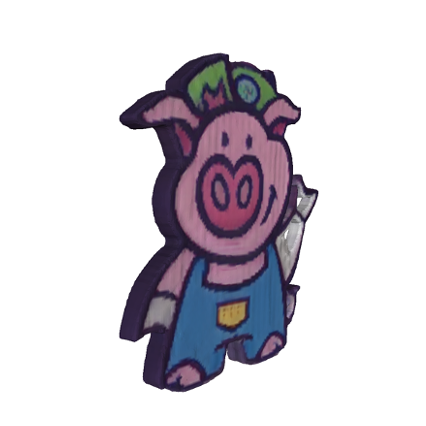}
        \includegraphics[width=\textwidth]{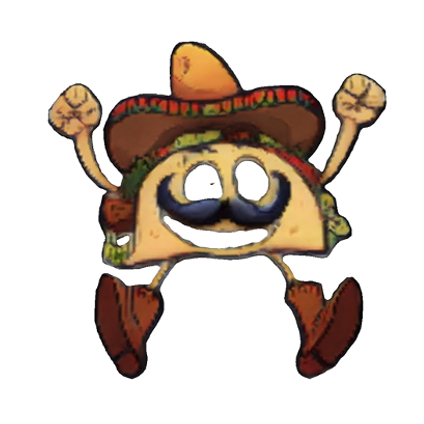}
        \caption{InstantMesh~\cite{xu2024instantmeshefficient3dmesh}}
    \end{subfigure}\hspace{-0.1em}
    \begin{subfigure}[b]{0.122\textwidth}
        \centering
        \includegraphics[width=\textwidth]{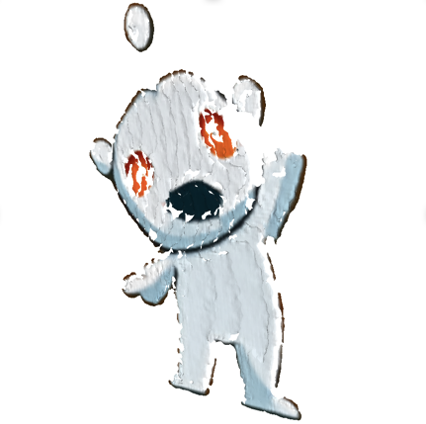}\\
        \includegraphics[width=\textwidth]{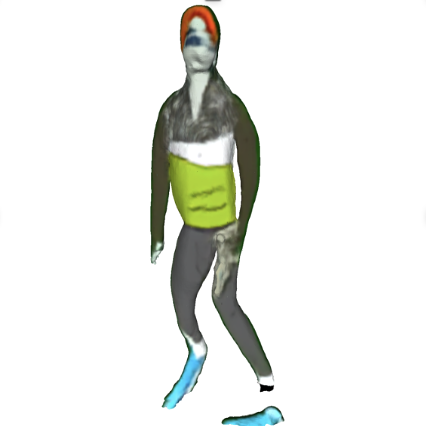}\\
        \includegraphics[width=\textwidth]{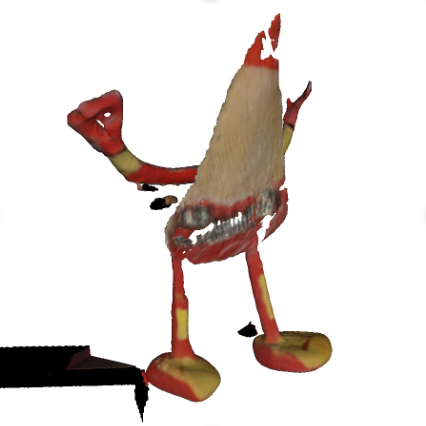}
        \includegraphics[width=\textwidth]{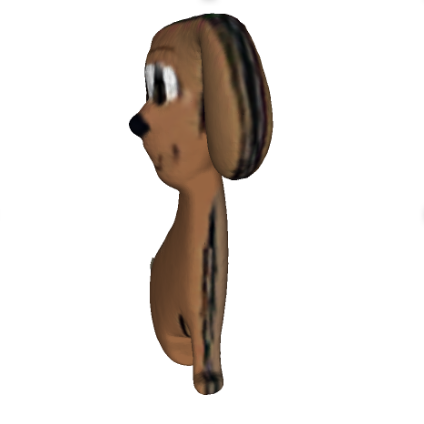}
        \includegraphics[width=\textwidth]{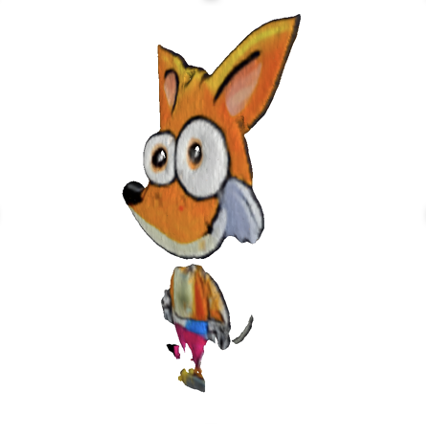}
        \includegraphics[width=\textwidth]{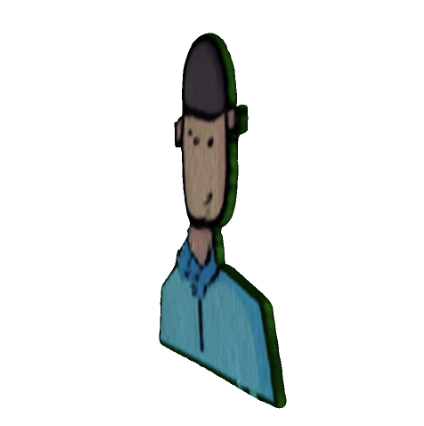}
        \includegraphics[width=\textwidth]{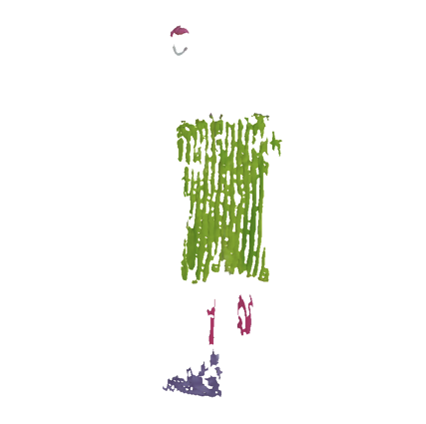}
        \includegraphics[width=\textwidth]{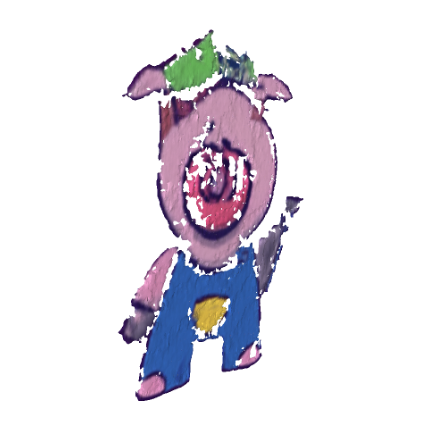}
        \includegraphics[width=\textwidth]{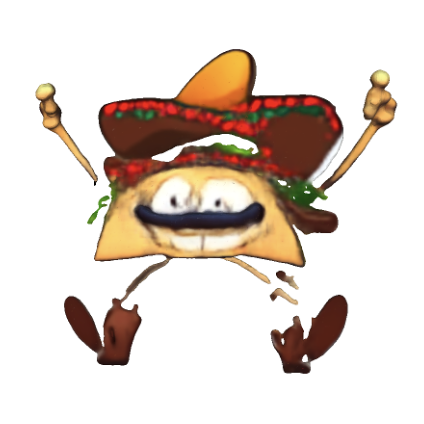}
        \caption{3DTopia-XL~\cite{chen20243dtopiaxlscalinghighquality3d}}
    \end{subfigure}\hspace{-0.1em}
    \begin{subfigure}[b]{0.122\textwidth}
        \centering
        \includegraphics[width=0.95\textwidth]{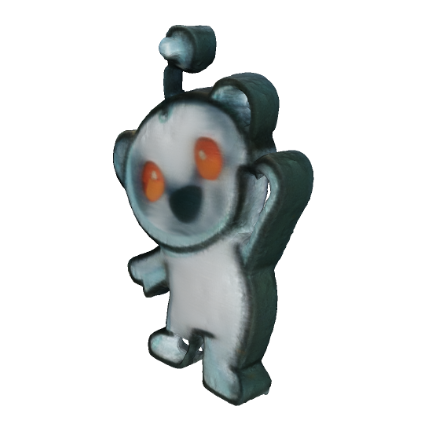}\\
        \includegraphics[width=\textwidth]{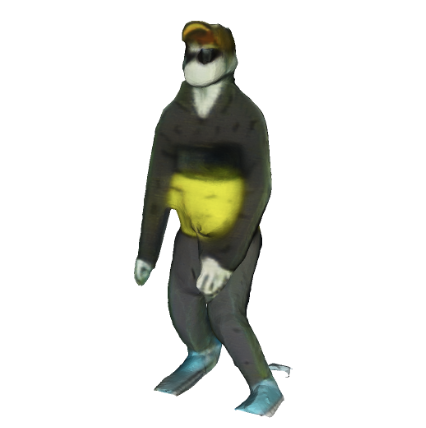}\\
        \includegraphics[width=\textwidth]{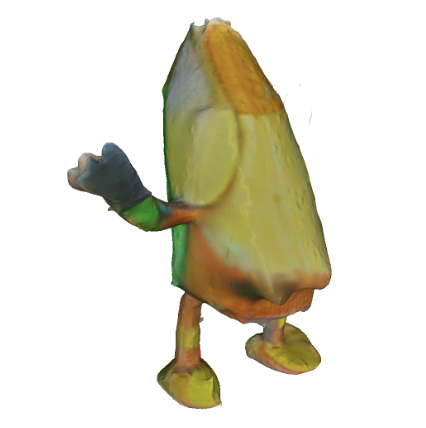}
        \includegraphics[width=\textwidth]{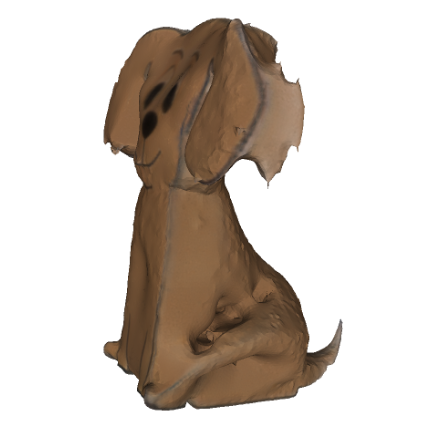}
        \includegraphics[width=\textwidth]{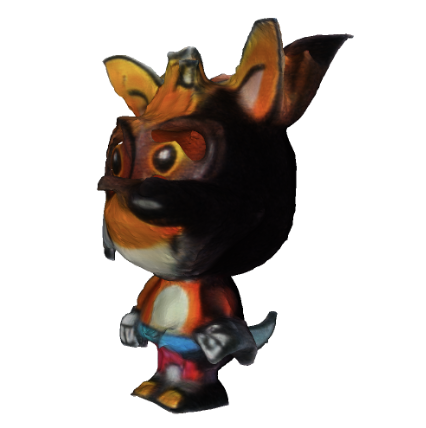}
        \includegraphics[width=\textwidth]{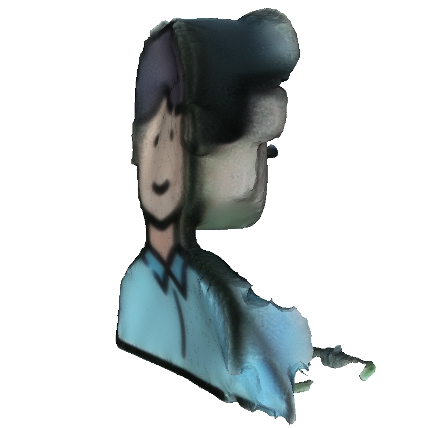}
        \includegraphics[width=\textwidth]{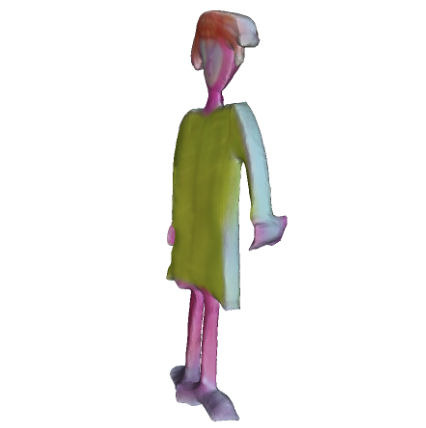}
        \includegraphics[width=\textwidth]{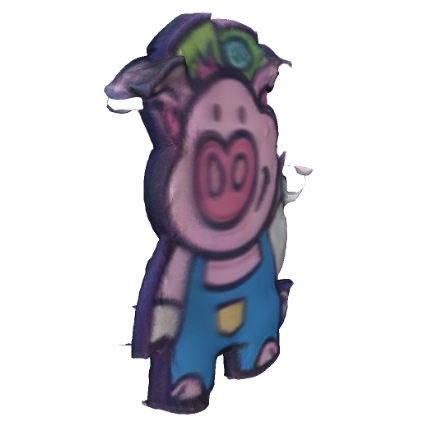}
        \includegraphics[width=\textwidth]{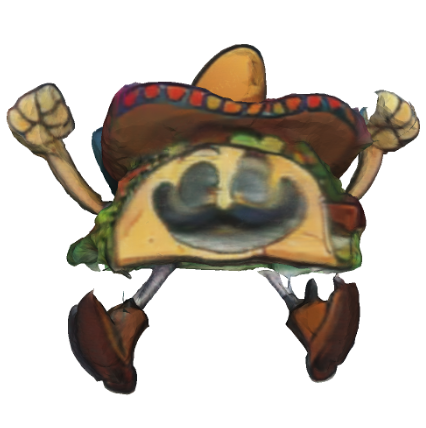}
        \caption{LGM~\cite{tang2024lgm}}
    \end{subfigure}\hspace{-0.1em}
    \begin{subfigure}[b]{0.122\textwidth}
        \centering
        \includegraphics[width=\textwidth]{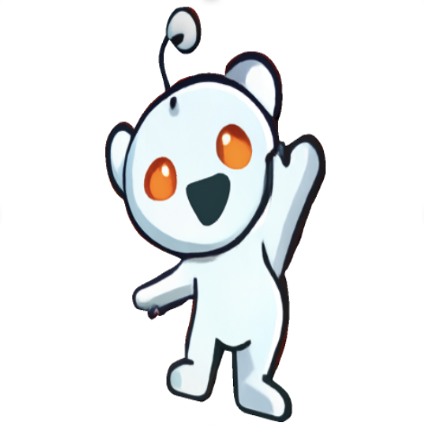}\\
        \includegraphics[width=\textwidth]{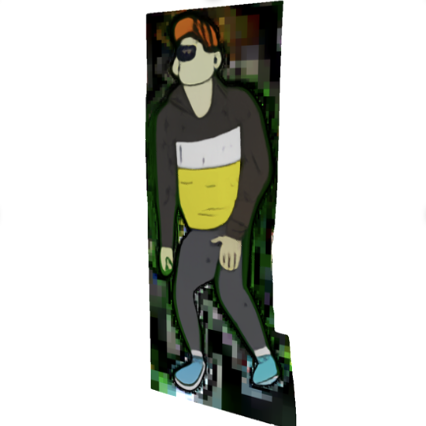}\\
        \includegraphics[width=\textwidth]{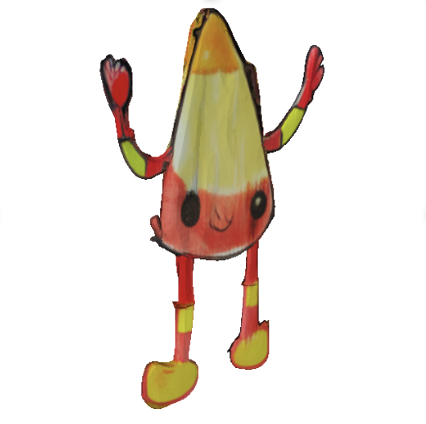}
        \includegraphics[width=\textwidth]{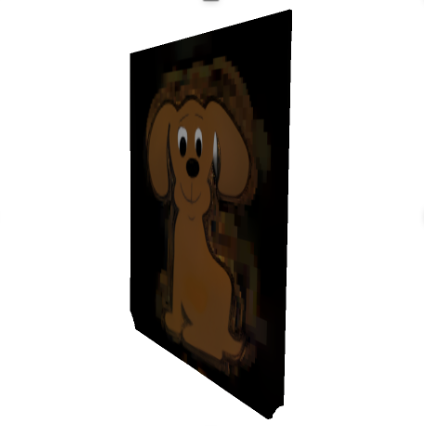}
        \includegraphics[width=\textwidth]{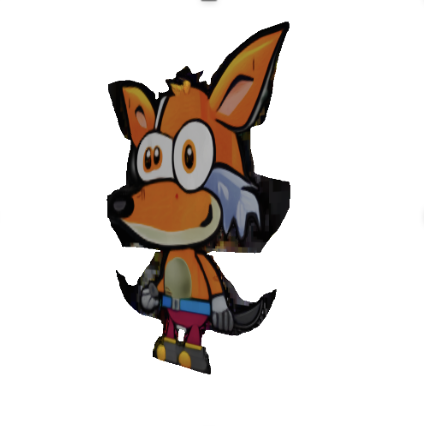}
        \includegraphics[width=\textwidth]{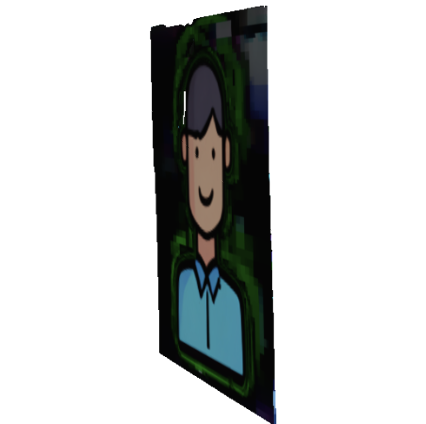}
        \includegraphics[width=\textwidth]{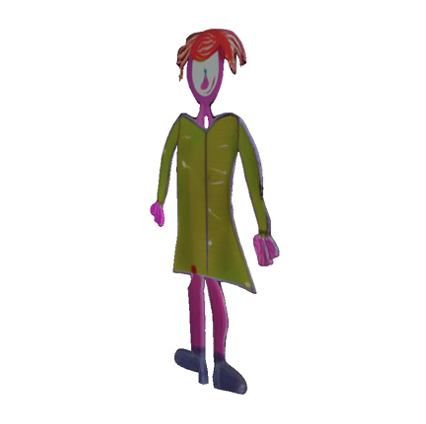}
        \includegraphics[width=\textwidth]{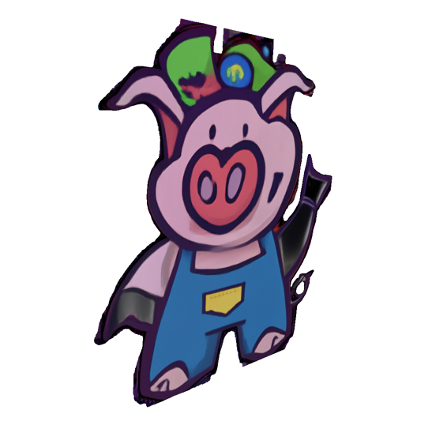}
        \includegraphics[width=\textwidth]{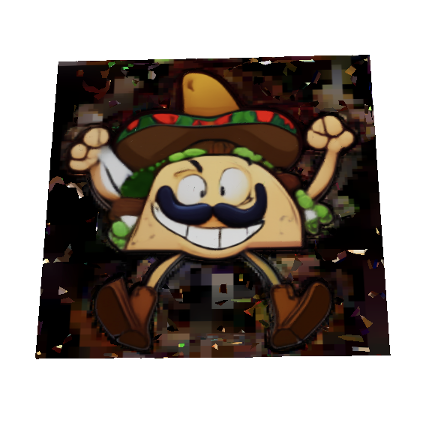}
        \caption{Trellis~\cite{xiang2024structured}}
    \end{subfigure}\hspace{-0.1em}
    \begin{subfigure}[b]{0.122\textwidth}
        \centering
        \includegraphics[width=\textwidth]{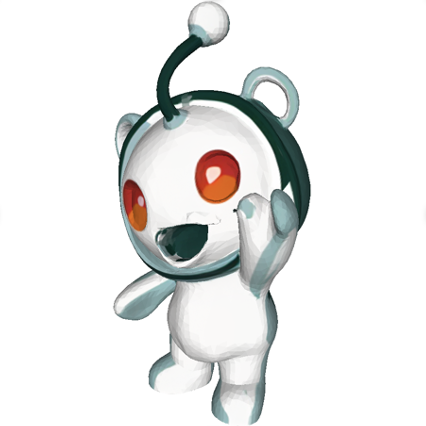}\\
        \includegraphics[width=\textwidth]{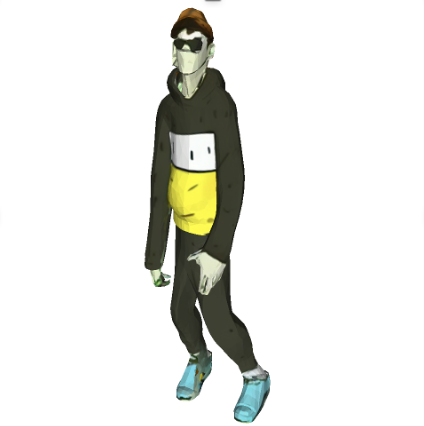}\\
        \includegraphics[width=\textwidth]{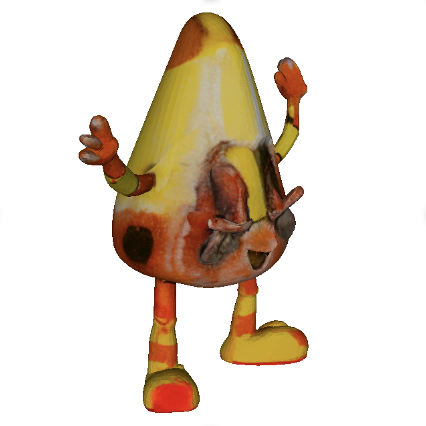}
        \includegraphics[width=\textwidth]{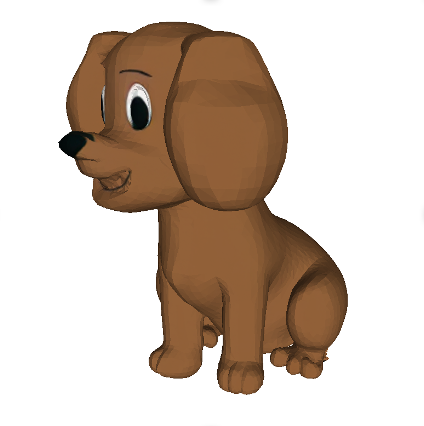}
        \includegraphics[width=\textwidth]{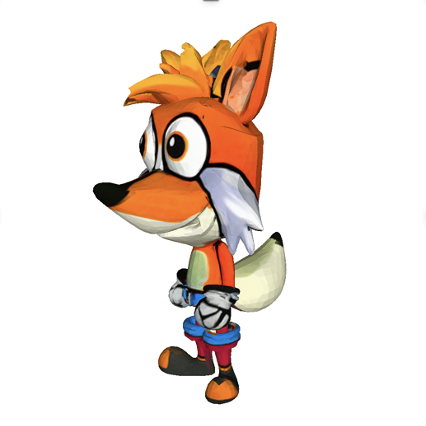}
        \includegraphics[width=\textwidth]{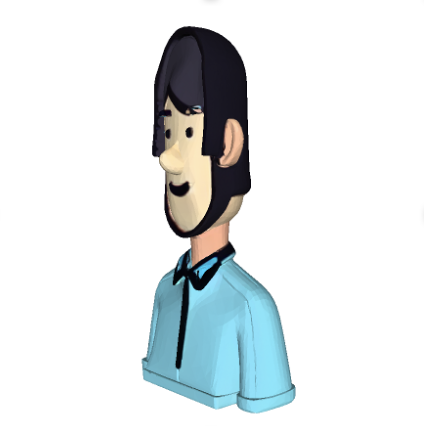}
        \includegraphics[width=\textwidth]{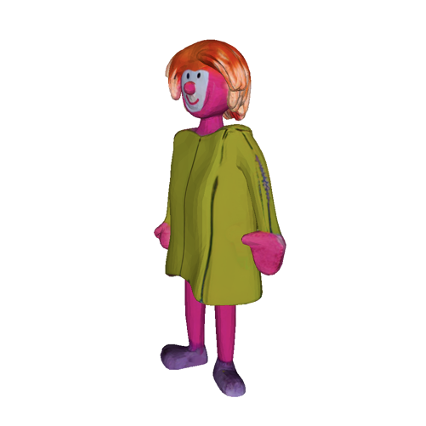}
        \includegraphics[width=\textwidth]{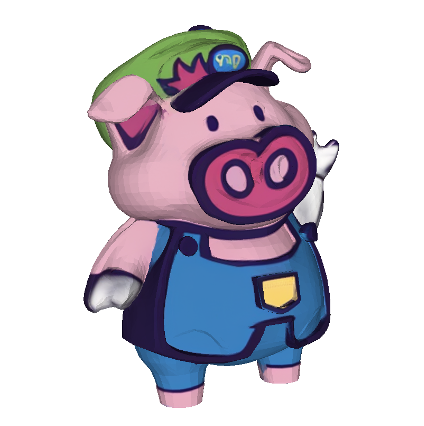}
        \includegraphics[width=\textwidth]{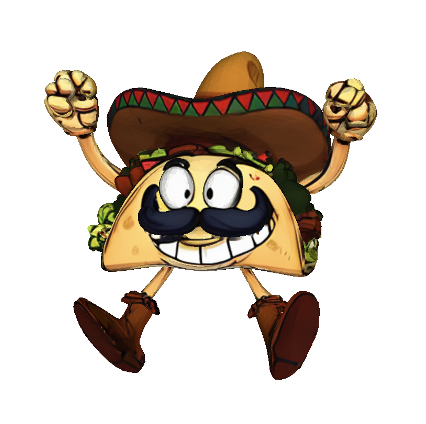}
        \caption{Ours}
    \end{subfigure}
    % \vspace{-0.1in}
    \caption{\textbf{Qualitative comparisons.} We combine our augmentation module with InstantMesh~\cite{xu2024instantmeshefficient3dmesh} and perform comparisons with state-of-the-art pre-trained image-to-3D methods Shap-E~\cite{jun2023shapegeneratingconditional3d}, LN3Diff~\cite{lan2024ln3diff}, InstantMesh~\cite{xu2024instantmeshefficient3dmesh}, 3DTopia-XL~\cite{chen20243dtopiaxlscalinghighquality3d}, LGM~\cite{tang2024lgm} and Trellis~\cite{xiang2024structured}. All the flat-shaped input images are from our curated dataset \textbf{\textit{Flat-2D}}. Our augmentation module can significantly improve the geometry quality of generated 3D assets.}
    % \vspace{-0.1in}
    \label{fig:qualitative_comparison}
\end{figure*}

% Finally, we feed $\hat{\mathbf{I}}$ to any image-to-3D models $\mathcal{R}$ to generate a high-quality 3D mesh $\mathbb{M} = \mathcal{R}(\hat{\mathbf{I}})$. 
% In this paper, we choose TRELLIS~\cite{xiang2024structured3dlatentsscalable} for all the experiments.

Finally, we use Trellis~\cite{xiang2024structured} $\mathcal{S}_G$ to generate a complete and reasonable 3D shape without texture based on the proxy image $\hat{\mathbf{I}}$.
To synthesize consistent appearance with $I$, we leverage Hunyuan2.0~\cite{zhao2025hunyuan3d20scalingdiffusion} $\mathcal{S}_T$ to bake a realistic texture under the guidance of the flat-colored input image and the generated shape. 
Thus, a textured mesh $\mathbb{M}$ can be generated by
\begin{equation}
    \mathbb{M} = \mathcal{S}_T(\mathcal{S}_G(\hat{\mathbf{I}}), I).
\end{equation}
By decoupling the generation procedure into two stages, \methodname achieves high-quality 3D asset generation with complete and reasonable geometry, as well as a realistic appearance consistent with the flat-colored image input.

\section{Experiments}

\subsection{Implementation Details}
\paragraph{Dataset.}
For benchmarking the performance of image-to-3D models on the flat images, we collect a new dataset \textbf{\datasetname} from public resources, including over $100$ high-quality flat images with various art styles, including hand drawings, cel painting, icons, cartoon-style designs, \etc.

\paragraph{Baselines.}
We adopt as baselines the most recent and related image-to-3D generative methods that utilize different generative paradigms, latent representations, and output formats, including Shap-E~\cite{jun2023shapegeneratingconditional3d}, LN3Diff~\cite{lan2024ln3diff}, InstantMesh~\cite{xu2024instantmeshefficient3dmesh}, 3DTopia-XL~\cite{chen20243dtopiaxlscalinghighquality3d}, LGM~\cite{tang2024lgm}, and Trellis~\cite{xiang2024structured}. 
For all the baselines, we use their official codes and pre-trained weights.

\subsection{Qualitative Comparisons}
We present qualitative visual comparisons on the $\textbf{\datasetname}$ dataset in Fig.~\ref{fig:qualitative_comparison}. 
\methodname is capable of generating high-quality 3D assets with complete and coherent geometry, as well as realistic appearances consistent with the flat input images. 
In contrast, all baseline methods exhibit significant degradation, often producing abnormally thin structures. 
This is mainly due to the distribution gap between flat input images and the training data of existing image-to-3D models, making it difficult for them to infer plausible 3D structures. 
\methodname mitigates this issue by enhancing the three-dimensional illusion of flat images by pre-trained 2D image diffusion models based on their structure features and semantic information, effectively eliminating the domain gap in shape synthesis.
\section{Conclusion}
We presented \methodname, a novel training-free 3D generation framework for \textbf{flat-colored} images which inherently lack a sense of three-dimensionality. 
\methodname enhances the three-dimensional illusion of the flat-color image input based on the structure features and semantic information to reduce the domain gap in image-to-3d generation by leveraging the prior from pretrained 2D image diffusion models.
We also introduced \datasetname, a new benchmark dataset for over 100 flat images spanning diverse art styles such as hand drawings, cel painting, icons, cartoon-style designs, \etc. 
Our method achieves state-of-the-art performance in generating complete and realistic 3D assets from flat-colored images.\\

\paragraph{\textbf{Acknowledgements.}}
This work was supported by NSF CAREER grant \#2143576.

{
    \small
    \bibliographystyle{ieeenat_fullname}
    \bibliography{main}
}

% % WARNING: do not forget to delete the supplementary pages from your submission 
% \input{sec/6_appendix}

\end{document}